\documentclass[11pt,a4paper]{article}
\usepackage{amsmath}
\usepackage[hyperref]{emnlp2020}
\usepackage{times}
\usepackage{latexsym}
\usepackage{amssymb}
\usepackage{breqn}

\usepackage{enumitem}
\usepackage{booktabs}
\usepackage{graphicx}
\usepackage{scalerel,xparse}

\usepackage{microtype}
\usepackage{arydshln}
\aclfinalcopy 

\makeatletter
\def\thickhline{%
  \noalign{\ifnum0=`}\fi\hrule \@height \thickarrayrulewidth \futurelet
   \reserved@a\@xthickhline}
\def\@xthickhline{\ifx\reserved@a\thickhline
               \vskip\doublerulesep
               \vskip-\thickarrayrulewidth
             \fi
      \ifnum0=`{\fi}}
\makeatother

\newlength{\thickarrayrulewidth}
\title{Transformer Based Multi-Source Domain Adaptation}

\author{Dustin Wright \and Isabelle Augenstein \\
  Dept. of Computer Science \\
  University of Copenhagen \\
  Denmark \\
  \texttt{\{dw|augenstein\}@di.ku.dk}}

\date{}

\begin{document}
\maketitle
\begin{abstract}
In practical machine learning settings, the data on which a model must make predictions often come from a different distribution than the data it was trained on. 
Here, we investigate the problem of \textit{unsupervised multi-source domain adaptation}, where a model is trained on labelled data from multiple source domains and must make predictions on a domain for which no labelled data has been seen. 
Prior work with CNNs and RNNs has demonstrated the benefit of mixture of experts, where the predictions of multiple domain expert classifiers are combined; as well as domain adversarial training, to induce a domain agnostic representation space. Inspired by this, we investigate how such methods can be effectively applied to large pretrained transformer models. 
We find that domain adversarial training has an effect on the learned representations of these models while having little effect on their performance, suggesting that large transformer-based models are already relatively robust across domains.
Additionally, we show that mixture of experts leads to significant performance improvements by comparing several variants of mixing functions, including one novel mixture based on attention. 
Finally, we demonstrate that the predictions of large pretrained transformer based domain experts are highly homogenous, making it challenging to learn effective functions for mixing their predictions.

\end{abstract}

\section{Introduction}
\begin{figure}[t]
  
  \centering
    \includegraphics[width=0.98\columnwidth]{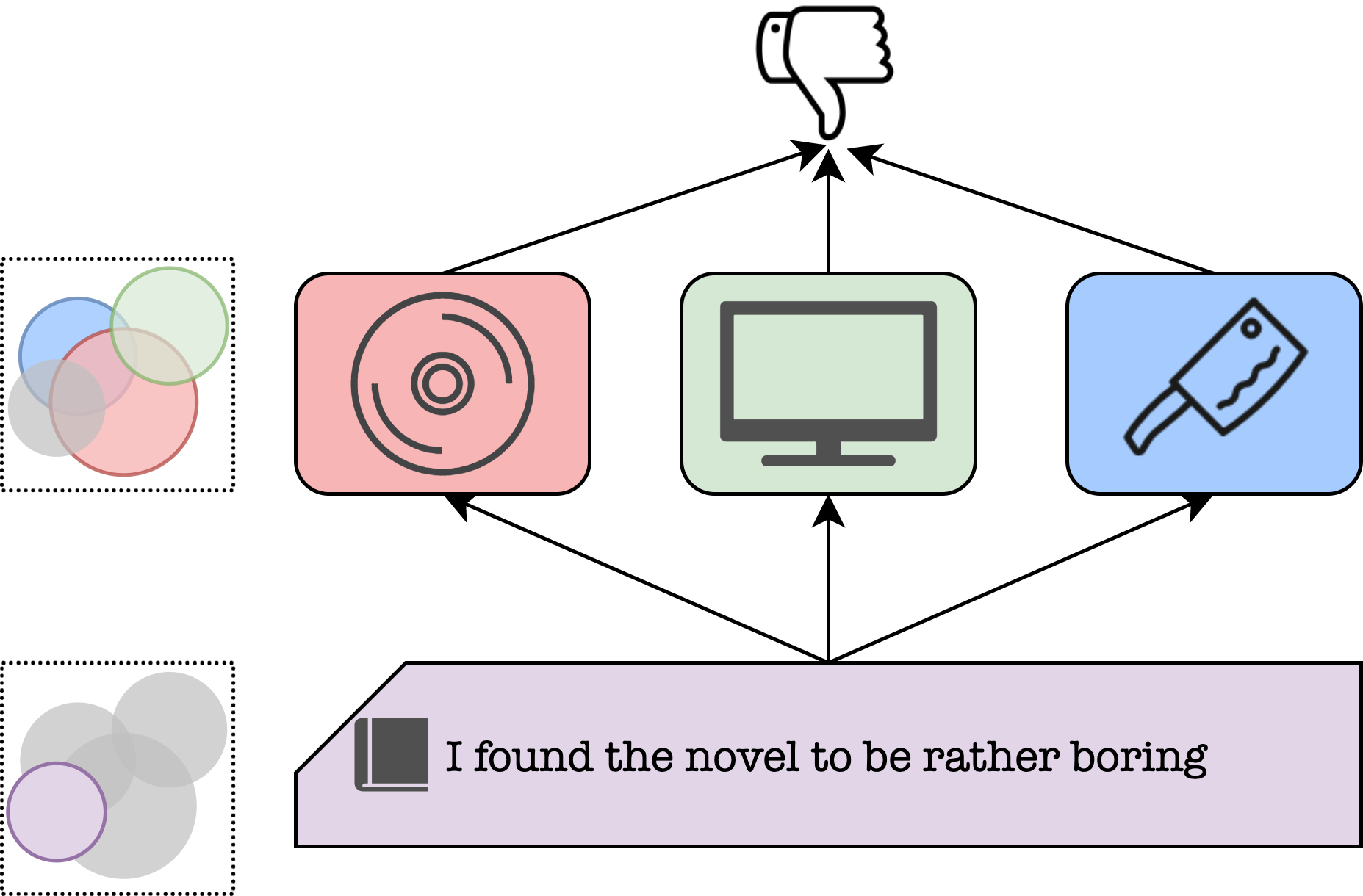}
    \caption{In multi-source domain adaptation, a model is trained on data drawn from multiple parts of the underlying distribution. At test time, the model must make predictions on data from a potentially non-overlapping part of the distribution.}
    \label{fig:msda}
\end{figure}
Machine learning practitioners are often faced with the problem of evolving test data, leading to mismatches in training and test set distributions.
As such, the problem of \textit{domain adaptation} is of particular interest to the natural language processing community in order to build models which are robust this shift in distribution. For example, a model may be trained to predict the sentiment of product reviews for DVDs, electronics, and kitchen goods, and must utilize this learned knowledge to predict the sentiment of a review about a book (\autoref{fig:msda}). This paper is concerned with this setting, namely \textit{unsupervised multi-source domain adaptation}.

Multi-source domain adaptation is a well studied problem in deep learning for natural language processing. Prominent techniques are generally based on data selection strategies and representation learning. For example, a popular representation learning method is to induce domain invariant representations using unsupervised target data and domain adversarial learning~\cite{ganin2015unsupervised}. Adding to this, mixture of experts techniques attempt to learn both domain specific and global shared representations and combine their predictions~\cite{guo2018multi,li2018s,ma2019domain}. These methods have been primarily studied using convolutional nets (CNNs) and recurrent nets (RNNs) trained from scratch, while the NLP community has recently begun to rely more and more on large pretrained transformer (LPX) models e.g. BERT~\cite{devlin2019bert}. 
To date there has been some preliminary investigation of how LPX models perform under domain shift in the single source-single target setting~\cite{ma2019domain,han2019unsupervised,rietzler2019adapt,gururangan2020don}. What is lacking is a study into the effects of and best ways to apply classic multi-source domain adaptation techniques with LPX models, which can give insight into possible avenues for improved application of these models in settings where there is domain shift.


Given this, we present a study into unsupervised multi-source domain adaptation techniques for large pretrained transformer models. 
Our main research question is: do mixture of experts and domain adversarial training offer any benefit when using LPX models? The answer to this is not immediately obvious, as such models have been shown to generalize quite well across domains and tasks while still learning representations which are not domain invariant. Therefore, we experiment with four mixture of experts models, including one novel technique based on attending to different domain experts; as well as domain adversarial training with gradient reversal. 
Surprisingly, we find that, while domain adversarial training helps the model learn more domain invariant representations, this does not always result in increased target task performance.
When using mixture of experts, we see significant gains on out of domain rumour detection, and some gains on out of domain sentiment analysis. Further analysis reveals that the classifiers learned by domain expert models are highly homogeneous, making it challenging to learn a better mixing function than simple averaging.



\section{Related Work}
Our primary focus is multi-source domain adaptation with LPX models. We first review domain adaptation in general, followed by studies into domain adaptation with LPX models.

\subsection{Domain Adaptation}
Domain adaptation approaches generally fall into three categories: \textit{supervised} approaches (e.g. \citet{daumeiii:2007:ACLMain,finkel-manning-2009-hierarchical-fixed,conf/cvpr/KulisSD11}), where both labels for the source and the target domain are available; \textit{semi-supervised} approaches (e.g. \citet{conf/cvpr/DonahueHRSD13,conf/cvpr/YaoPNLM15}), where labels for the source and a small set of labels for the target domain are provided; and lastly \textit{unsupervised} approaches (e.g. \citet{blitzer-etal-2006-domain-fixed,ganin2015unsupervised,conf/aaai/SunFS16,conf/icml/LiptonWS18}), where only labels for the source domain are given. Since the focus of this paper is the latter, we restrict our discussion to unsupervised approaches. A more complete recent review of unsupervised domain adaptation approaches is given in \citet{kouw2019review}.

A popular approach to unsupervised domain adaptation is to induce representations which are invariant to the shift in distribution between source and target data. For deep networks, this can be accomplished via domain adversarial training using a simple gradient reversal trick~\cite{ganin2015unsupervised}. This has been shown to work in the multi-source domain adaptation setting too~\cite{li2018s}. Other popular representation learning methods include minimizing the covariance between source and target features~\cite{conf/aaai/SunFS16} and using maximum-mean discrepancy between the marginal distribution of source and target features as an adversarial objective~\cite{guo2018multi}.

Mixture of experts has also been shown to be effective for multi-source domain adaptation. \citet{kim2017domain} use 
attention to combine the predictions of domain experts. \citet{guo2018multi} propose learning a mixture of experts using a point to set metric, which combines the posteriors of models trained on individual domains. Our work attempts to build on this to study how multi-source domain adaptation can be improved with LPX models.

\subsection{Transformer Based Domain Adaptation}
There are a handful of studies which investigate how LPX models can be improved in the presence of domain shift. These methods tend to focus on the data and training objectives for single-source single-target unsupervised domain adaptation. The work of \citet{ma2019domain} shows that curriculum learning based on the similarity of target data to source data improves the performance of BERT on out of domain natural language inference. Additionally, \citet{han2019unsupervised} demonstrate that domain adaptive fine-tuning with the masked language modeling objective of BERT leads to improved performance on domain adaptation for sequence labelling. \citet{rietzler2019adapt} offer similar evidence for task adaptive fine-tuning on aspect based sentiment analysis. \citet{gururangan2020don} take this further, showing that significant gains in performance are yielded when progressively fine-tuning on in domain data, followed by task data, using the masked language modeling objective of RobERTa. Finally, \citet{lin2020does} explore whether domain adversarial training with BERT would improve performance for clinical negation detection, finding that the best performing method is a plain BERT model, giving some evidence that perhaps well-studied domain adaptation methods may not be applicable to LPX models.

What has not been studied, to the best of our knowledge, is the impact of domain adversarial training via gradient reversal on LPX models on natural language processing tasks, as well as if mixture of experts techniques can be beneficial. As these methods have historically benefited deep 
models for domain adaptation, we explore their effect when applied to LPX models in this work. 

\section{Methods}
\begin{figure}[t]
  
  \centering
    \includegraphics[width=\columnwidth]{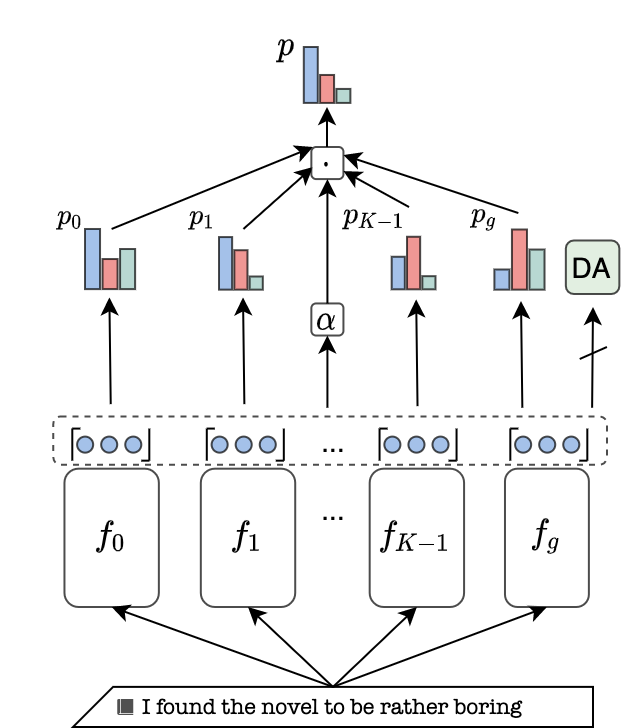}
    \caption{The overall approach tested in this work. A sample is input to a set of expert and one shared LPX model as described in \S\ref{sec:modeling}. The output probabilities of these models are then combined using an attention parameter alpha (\S\ref{sec:avg}, \S\ref{sec:fta}, \S\ref{sec:dc}, \S\ref{sec:attention}). In addition, a global model $f_g$ learns domain invariant representations via a classifier \texttt{DA} with gradient reversal (indicated by the slash, see \S\ref{sec:da_method}).}
    \label{fig:check-worthy-examples}
\end{figure}
This work is motivated by previous research on domain adversarial training and mixture of domain experts for domain adaptation. In this, the data consists of 
$K$ source domains $\mathcal{S}$ and a target domain $\mathcal{T}$. The source domains consist of labelled datasets $D_{s}, s \in \{1,...,K\}$ and the target domain consists only of unlabelled data $U_{t}$. The goal is to learn a classifier $f$, which generalizes well to $\mathcal{T}$ using only the labelled data from $\mathcal{S}$ and optionally unlabelled data from $\mathcal{T}$. We consider a base network $f_{z}, z \in \mathcal{S} \cup \{g\}$ corresponding to either a domain specific network or a global shared network. These $f_{z}$ networks are initialized using LPX models, in particular DistilBert~\cite{sanh2019distilbert}. 

\subsection{Mixture of Experts Techniques}\label{sec:modeling}
We study four different mixture of expert techniques: simple averaging, fine-tuned averaging, attention with a domain classifier, and a novel sample-wise attention mechanism based on transformer attention~\cite{vaswani2017attention}. Prior work reports that utilizing mixtures of domain experts and shared classifiers leads to improved performance when having access to multiple source domains~\cite{guo2018multi,li2018s}. Given this, we investigate if mixture of experts can have any benefit when using LPX models. 

Formally, for a setting with $K$ domains, we have set of $K$ different LPX models $f_{k}, k \in \{0...K-1\}$ corresponding to each domain. There is also an additional LPX model $f_{g}$ corresponding to a global shared model. The output predictions of these models are $p_{k}, k \in \{0...K-1\}$ and $p_{g}$, respectively. Since the problems we are concerned with are binary classification, these are single values in the range $(0,1)$. The final output probability is calculated as a weighted combination of a set of domain expert probabilities $\bar{\mathcal{K}} \subseteq \mathcal{S}$ and the probability from the global shared model. Four methods are used for calculating the weighting. 

\subsubsection{Averaging}
\label{sec:avg}
The first method is a simple averaging of the predictions of domain specific and shared classifiers. The final output of the model is
\begin{equation}
    p_A(x,\bar{\mathcal{K}}) = \frac{1}{|\bar{\mathcal{K}}|+1}\sum_{k \in \bar{\mathcal{K}}}p_{k}(x) + p_{g}(x)
\end{equation}

\subsubsection{Fine Tuned Averaging}
\label{sec:fta}
As an extension to simple averaging, we fine tune the weight given to each of the domain experts and global shared model. This is performed via randomized grid search evaluated on validation data, after the models have been trained. A random integer between zero and ten is generated for each of the models, which is then normalized to a set of probabilities $\alpha_{F}$. The final output probability is then given as follows.

\begin{equation}
    p_F(x) = \sum_{k \in \bar{\mathcal{K}}}p_k(x) * \alpha_{F}^{(k)}(x) + p_g(x) * \alpha_{F}^{(g)}(x)
\end{equation}


\subsubsection{Domain Classifier}
\label{sec:dc}
It was recently shown that curriculum learning using a domain classifier can lead to improved performance for single-source domain adaptation~\cite{ma2019domain} when using LPX models. Inspired by this, we experiment with using a domain classifier as a way to attend to the predictions of domain expert models. First, a domain classifier $f_C$ is trained to predict the domain of an input sample $x$ given $\mathbf{r}_{g} \in \mathbb{R}^{d}$, the representation of the \texttt{[CLS]} token at the output of a LPX model. From the classifier, a vector $\alpha_{C}$ is produced with the probabilities that a sample belongs to each source domain. 
\begin{equation}
    \alpha_{C} = f_{C}(x) = \text{softmax}(\mathbf{W}_C\mathbf{r}_{g} + b_{C})
\end{equation}
where $\mathbf{W}_{C} \in \mathbb{R}^{d \times K}$ and $b_{C} \in \mathbb{R}^{K}$. The domain classifier is trained before the end-task network and is held static throughout training on the end-task. For this, a set of domain experts $f_{k}$ are trained and their predictions combined through a weighted sum of the attention vector $\alpha_{C}$.
\begin{equation}
    p_C(x) = \sum_{k \in S}p_k(x) * \alpha_C^{(k)}(x)
\end{equation}
where the superscript $(k)$ indexes into the $\alpha_C$ vector. Note that in this case we only use domain experts and not a global shared model. In addition, the probability is always calculated with respect to each source domain.

\subsubsection{Attention Model}
\label{sec:attention}
Finally, a novel parameterized attention model is learned which attends to different domains based on the input sample. The attention method is based on the scaled dot product attention applied in transformer models~\cite{vaswani2017attention}, where a global shared model acts as a query network attending to each of the expert and shared models. As such, a shared model $f_{g}$ produces a vector $\mathbf{r}_{g} \in \mathbb{R}^{d}$, and each domain expert produces a vector $\mathbf{r}_{k} \in \mathbb{R}^{d}$. First, for an input sample $x$, a probability for the end task is obtained from the classifier of each model yielding probabilities $p_{g}$ and $p_{k}, k \in {0...K-1}$.
An attention vector $\alpha_{X}$ is then obtained via the following transformations.
\begin{equation}
    \mathbf{q} = \mathbf{g}\mathbf{Q}^{T}
\end{equation}
\begin{equation}
    \mathbf{k} = \begin{bmatrix}
           \mathbf{r}_{1} \\
           \vdots \\
           \mathbf{r}_{K} \\
           \mathbf{r}_{g}
         \end{bmatrix} \mathbf{K}^{T}
\end{equation}
\begin{equation}
    \alpha_{X} = \text{softmax}(\mathbf{q}\mathbf{k}^{T})
\end{equation}
where $\mathbf{Q} \in \mathbb{R}^{d \times d}$ and $\mathbf{K} \in \mathbb{R}^{d \times d}$. The attention vector $\alpha_{X}$ then attends to the individual predictions of each domain expert and the global shared model. 
\begin{equation}
    p_X(x,\bar{\mathcal{K}}) = \sum_{k \in \bar{\mathcal{K}}}p_k(x) * \alpha_{X}^{(k)}(x) + p_g(x) * \alpha_{X}^{(g)}(x)
\end{equation}

To ensure that each model is trained as a domain specific expert, a similar training procedure to that of \citealt{guo2018multi} is utilized, described in \S\ref{sec:training}.

\subsection{Domain Adversarial Training}
\label{sec:da_method}
The method of domain adversarial adaptation we investigate here is the well-studied technique described in~\citet{ganin2015unsupervised}. It has been shown to benefit both convolutional nets and recurrent nets on NLP problems~\cite{li2018s,gui2017part}, so is a prime candidate to study in the context of LPX models. Additionally, some preliminary evidence indicates that adversarial training might improve LPX generalizability for single-source domain adaptation~\cite{ma2019domain}. 

To learn domain invariant representations, we train a model such that the learned representations maximally confuse a domain classifier $f_d$. This is accomplished through a min-max objective between the domain classifier parameters $\theta_{D}$ and the parameters $\theta_{G}$ of an encoder $f_g$. The objective can then be described as follows.
\begin{equation}
    \mathcal{L}_D = \max_{\theta_{D}}\min_{\theta_{G}}-d\log  f_{d}(f_{g}(x))
\end{equation}
where $d$ is the domain of input sample $x$. The effect of this is to improve the ability of the classifier to determine the domain of an instance, while encouraging the model to generate maximally confusing representations via minimizing the negative loss. In practice, this is accomplished by training the model using standard cross entropy loss, but reversing the gradients of the loss with respect to the model parameters $\theta_{G}$.

\subsection{Training}
\label{sec:training}
Our training procedure follows a multi-task learning setup in which the data from a single batch comes from a single domain. Domains are thus shuffled on each round of training and the model is optimized for a particular domain on each batch. 

For the attention based (\S\ref{sec:attention}) and averaging (\S\ref{sec:avg}) models we adopt a similar training algorithm to~\citet{guo2018multi}. For each batch of training, a meta-target $t$ is selected from among the source domains, with the rest of the domains treated as meta-sources $\mathcal{S}' \in \mathcal{S} \setminus \{t\}$. Two losses are then calculated. The first is with respect to all of the meta-sources, where the attention vector is calculated for only those domains. For target labels $y_{i}$ and a batch of size $N$ with samples from a single domain, this is given as follows.
\begin{equation}
    \mathcal{L}_{s} = -\frac{1}{N}\sum_{i}y_{i}\log p_{X}(x, \mathcal{S}')
\end{equation}
The same procedure is followed for the averaging model $p_{A}$. The purpose is to encourage the model to learn attention vectors for out of domain data, thus why the meta-target is excluded from the calculation. 

The second loss is with respect to the meta-target, where the cross-entropy loss is calculated directly for the domain expert network of the meta-target.
\begin{equation}
    \mathcal{L}_{t} = -\frac{1}{N}\sum_{i}y_{i}\log p_t(x)
\end{equation}
This allows each model to become a domain expert through strong supervision. The final loss of the network is a combination of the three losses described previously, with $\lambda$ and $\gamma$  hyperparameters controlling the weight of each loss.
\begin{equation}
    \mathcal{L} = \lambda \mathcal{L}_s + (1 - \lambda) \mathcal{L}_t + \gamma \mathcal{L}_D
\end{equation}

For the domain classifier (\S\ref{sec:dc}) and fine-tuned averaging (\S\ref{sec:fta}), the individual LPX models are optimized directly with no auxiliary mixture of experts objective. In addition, we experiment with training the simple averaging model directly.

\section{Experiments and Results}

\begin{table*}[t!]
    \centering
    \fontsize{10}{10}\selectfont
    \begin{tabular}{l c c c c c | c c c c c c c }
    \toprule 
     Method &\multicolumn{5}{c}{Sentiment Analysis (Accuracy)}&\multicolumn{6}{c}{Rumour Detection (F1)}\\
    \cmidrule(lr){2-6}
    \cmidrule(lr){7-12}
     & D & E & K & B & macroA & CH & F & GW & OS & S & $\mu$F1\\
    \midrule 
         \citealt{li2018s} &77.9 &80.9 &80.9 &77.1 & 79.2 & - & - & - & - & - & -\\
      \citealt{guo2018multi} &87.7 &89.5 & 90.5& 87.9 &88.9 & - & - & - & - & - & -\\
      \citealt{zubiaga2017exploiting}&- &- &- &- & - & 63.6& \textbf{46.5}& 70.4& 69.0& 61.2& 60.7\\
    \midrule
        Basic & 89.1& 89.8& 90.1& 89.3& 89.5& 66.1& 44.7& 71.9& 61.0& 63.3& 62.3\\
        \midrule
        Adv-6 & 88.3& 89.7& 90.0& 89.0& 89.3 & 65.8& 42.0& 66.6& 61.7& 63.2& 61.4\\
        Adv-3 & 89.0& 89.9& 90.3& 89.0& 89.6& 65.7& 43.2& 72.3& 60.4& 62.1& 61.7\\
        \midrule
        Independent-Avg & 88.9& \textbf{90.6}& 90.4& \textbf{90.0}& \textbf{90.0}& 66.1& 45.6& 71.7& 59.4& 63.5& 62.2\\
        Independent-Ft & 88.9& 90.3& \textbf{90.8}& \textbf{90.0}& \textbf{90.0}& 65.9& 45.7& 72.2& 59.3& 62.4& 61.9\\
        MoE-Avg & \textbf{89.3}& 89.9& 90.5& 89.9& 89.9& \textbf{67.9}& 45.4& \textbf{74.5}& 62.6& \textbf{64.7}& \textbf{64.1}\\
        MoE-Att & 88.6& 90.0& 90.4& 89.6& 89.6& 65.9& 42.3& 72.5& 61.2& 63.3& 62.2\\
        MoE-Att-Adv-6 & 87.8& 89.0 & 90.5& 88.3& 88.9& 66.0& 40.7& 69.0& 63.8& 63.7& 61.8\\
        MoE-Att-Adv-3 & 88.6& 89.1& 90.4& 88.9& 89.2& 65.6& 42.7& 73.4& 60.9& 61.0& 61.8\\
        MoE-DC & 87.8& 89.2& 90.2& 87.9& 88.8& 66.5& 40.6& 70.5& \textbf{70.8}& 62.8& 63.8\\
    \bottomrule 

    \end{tabular}
    \caption{Experiments for sentiment analysis in (D)VD, (E)lectronics, (K)itchen and housewares, and (B)ooks domains and rumour detection for different events ((C)harlie(H)ebdo, (F)erguson, (G)erman(W)ings, (O)ttawa(S)hooting, and (S)ydneySiege) using leave-one-out cross validation. Results are averaged across 5 random seeds. The results for sentiments analysis are in terms of accuracy and the results for rumour detection are in terms of F1.}
    \label{tab:sentiment_results}
\end{table*}
We focus our experiments on text classification problems with data from multiple domains. To this end, we experiment with sentiment analysis from Amazon product reviews and rumour detection from tweets. For both tasks, we perform cross-validation on each domain, holding out a single domain for testing and training on the remaining domains, allowing a comparison of each method on how well they perform under domain shift. The code to reproduce all of the experiments in this paper can be found here\footnote{\url{https://github.com/copenlu/xformer-multi-source-domain-adaptation}}.

\paragraph{Sentiment Analysis Data} The data used for sentiment analysis come from the legacy Amazon Product Review dataset~\cite{blitzer2007biographies}. This dataset consists of 8,000 total tweets from four product categories: books, DVDs, electronics, and kitchen and housewares. Each domain contains 1,000 positive and 1,000 negative reviews. In addition, each domain has associated unlabelled data. Following previous work we focus on the transductive setting~\cite{guo2018multi,ziser2017neural} where we use the same 2,000 out of domain tweets as unlabelled data for training the domain adversarial models.  This data has been well studied in the context of domain adaptation, making for easy comparison with previous work.

\paragraph{Rumour Detection Data} The data used for rumour detection come from the PHEME dataset of rumourous tweets~\cite{zubiaga2016analysing}. There are a total of 5,802 annotated tweets from 5 different events labelled as rumourous or non-rumourous (1,972 rumours, 3,830 non-rumours). Methods which have been shown to work well on this data include context-aware classifiers \cite{zubiaga2017exploiting} and positive-unlabelled learning \cite{wright2020claim}. Again, we use this data in the transductive setting when testing domain adversarial training. 

\subsection{Baselines}

\paragraph{What's in a Domain?}
We use the model from~\citet{li2018s} as a baseline for sentiment analysis. This model consists of a set of domain experts and one general CNN, and is trained with a domain adversarial auxiliary objective.

\paragraph{Mixture of Experts}
Additionally, we present the results from~\citet{guo2018multi} representing the most recent state of the art on the Amazon reviews dataset. Their method consists of domain expert classifiers trained on top of a shared encoder, with predictions being combined via a novel metric which considers the distance between the mean representations of target data and source data. 

\paragraph{\citealt{zubiaga2017exploiting}} Though not a domain adaptation technique, we include the results from \citealt{zubiaga2017exploiting} on rumour detection to show the current state of the art performance on this task. The model is a CRF, which utilizes a combination of content and social features acting on a timeline of tweets.
\begin{figure*}[t]
  
  \centering
    \includegraphics[width=0.9\textwidth]{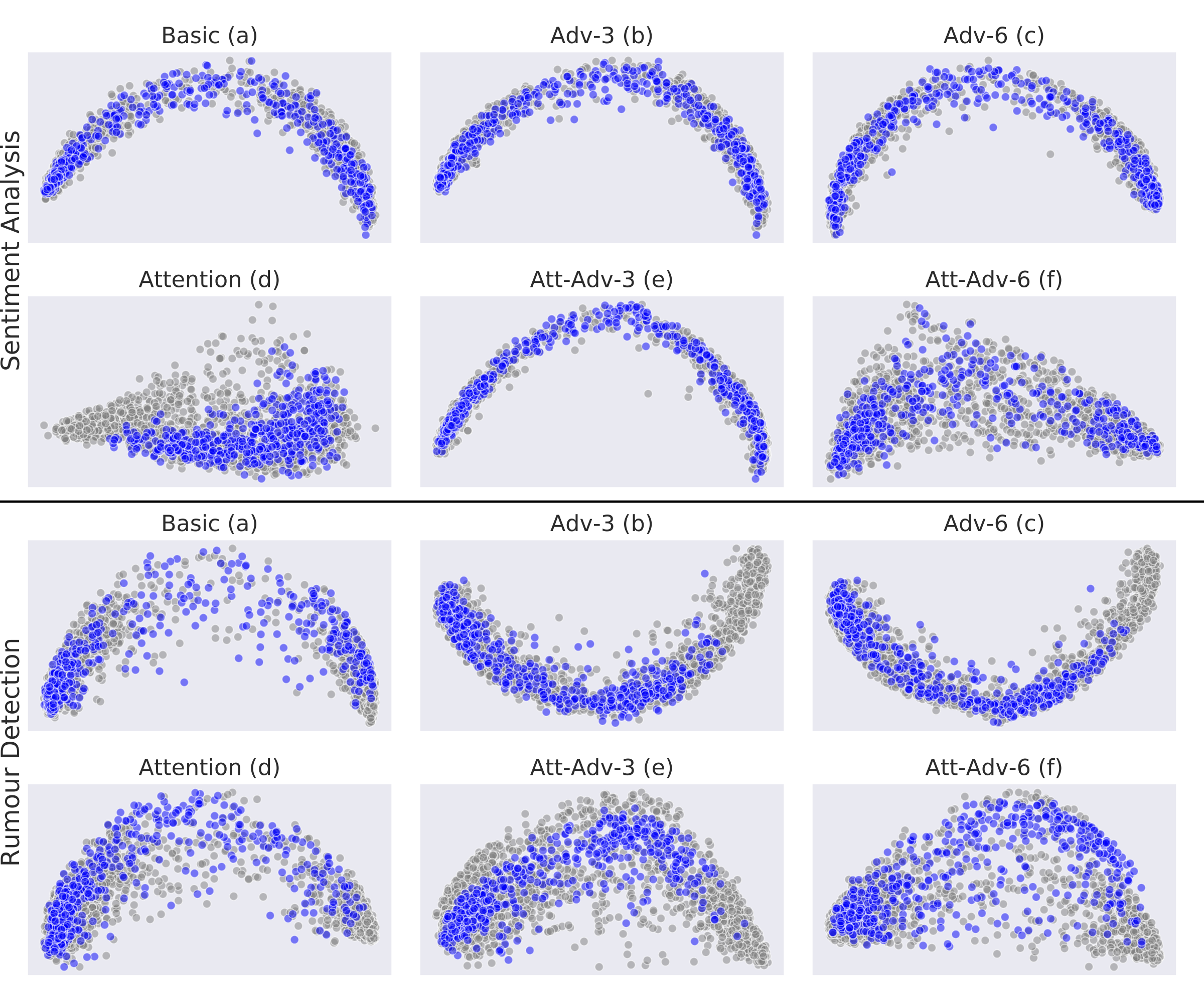}
    \caption{Final layer DistilBert embeddings for 500 randomly selected examples from each split for each tested model for sentiment data (top two rows) and rumour detection (bottom two rows). The blue points are out of domain data (in this case from Kitchen and Housewares for sentiment analysis and Sydney Siege for rumour detection) and the gray points are in domain data. }
    \label{fig:all_reps}
\end{figure*}

\subsection{Model Variants}
A variety of models are tested in this work. Therefore, each model is referred to by the following.

\paragraph{Basic} Basic DistilBert with a single classification layer at the output. 

\paragraph{Adv-$X$} DistilBert with domain adversarial supervision applied at the $X$'th layer (\S\ref{sec:da_method}).

\paragraph{Independent-Avg} DistilBert mixture of experts averaged but trained individually (not with the algorithm described in \S\ref{sec:training}).

\paragraph{Independent-FT} DistilBert mixture of experts averaged with mixing attention fine tuned after training (\S\ref{sec:fta}), trained individually.

\paragraph{MoE-Avg} DistilBert mixture of experts using averaging (\S\ref{sec:avg}).

\paragraph{MoE-Att} DistilBert mixture of experts using our novel attention based technique (\S\ref{sec:attention}).

\paragraph{MoE-Att-Adv-$X$} DistilBert mixture of experts using attention and domain adversarial supervision applied at the $X$'th layer.

\paragraph{MoE-DC} DistilBert mixture of experts using a domain classifier for attention (\S\ref{sec:dc}).

\subsection{Results}

  

Our results are given in \autoref{tab:sentiment_results}. Similar to the findings of \citet{lin2020does} on clinical negation, we see little overall difference in performance from both the individual model and the mixture of experts model when using domain adversarial training on sentiment analysis. For the base model, there is a slight improvement when domain adversarial supervision is applied at a lower layer of the model, but a drop when applied at a higher level. Additionally, mixture of experts provides some benefit, especially using the simpler methods such as averaging.


For rumour detection, again we see little performance change from using domain adversarial training, with a slight drop when supervision is applied at either layer. The mixture of experts methods overall perform better than single model methods, suggesting that mixing domain experts is still effective when using large pretrained transformer models. In this case, the best mixture of experts methods are simple averaging and static grid search for mixing weights, indicating the difficulty in learning an effective way to mix the predictions of domain experts. We elaborate on our findings further in \S\ref{sec:discussion}. Additional experiments on domain adversarial training using Bert can be found in \autoref{tab:bert_appendix_results} in \S\ref{sec:bert_appendix}, where we similarly find that domain adversarial training leads to a drop in performance on both datasets.

\section{Discussion}
\label{sec:discussion}
We now discuss our initial research questions in light of the results we obtained, and provide explanations for the observed behavior.
  
\subsection{What is the Effect of Domain Adversarial Training?}
We present PCA plots of the representations learned by different models in \autoref{fig:all_reps}. These are the final layer representations of 500 randomly sampled points for each split of the data. In the ideal case, the representations for out of domain samples would be indistinguishable from the representations for in domain data.

In the case of basic DistilBert, we see a slight change in the learned representations of the domain adversarial models versus the basic model (\autoref{fig:all_reps} top half, a-c) for sentiment analysis. When the attention based mixture of experts model is used, the representations of out of domain data cluster in one region of the representation space (d). With the application of adversarial supervision, the model learns representations which are more domain agnostic. Supervision applied at layer 6 of DistilBert (plot f) yields a representation space similar to the version without domain adversarial supervision. Interestingly, the representation space of the model with supervision at layer 3 (plot e) yields representations similar to the basic classifier. This gives some potential explanation as to the similar performance of this model to the basic classifier on this split (kitchen and housewares). Overall, domain adversarial supervision has some effect on performance, leading to gains in both the basic classifier and the mixture of experts model for this split. Additionally, there are minor improvements overall for the basic case, and a minor drop in performance with the mixture of experts model.



The effect of domain adversarial training is more pronounced on the rumour detection data for the basic model (\autoref{fig:all_reps} bottom half, a), where the representations exhibit somewhat less variance when domain adversarial supervision is applied. Surprisingly, this leads to a slight drop in performance for the split of the data depicted here (Sydney Siege). For the attention based model, the variant without domain adversarial supervision (d) already learns a somewhat domain agnostic representation. The model with domain adversarial supervision at layer 6 (f) furthers this, and the classifier learned from these representations perform better on this split of the data. Ultimately, the best performing models for rumour detection do not use domain supervision, and the effect on performance on the individual splits are mixed, suggesting that domain adversarial supervision can potentially help, but not in all cases.


\subsection{Is Mixture of Experts Useful with LPX Models?}
\label{sec:moe_discussion}
\begin{figure}
  
  \centering
    \includegraphics[width=\columnwidth]{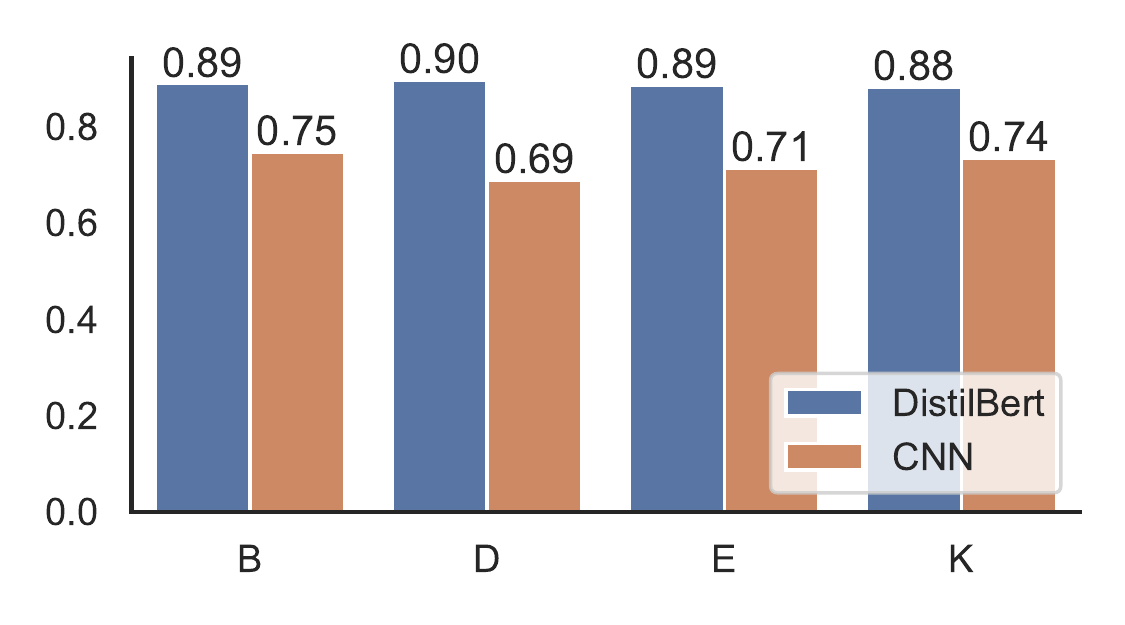}
    \caption{Comparison of agreement (Krippendorff's alpha) between domain expert models when the models are either DistilBert or a CNN. Predictions are made on unseen test data by each domain expert, and agreement is measured between their predictions ((B)ooks, (D)VD, (E)lectronics, and (K)itchen). The overall agreement between the DistilBert experts is greater than the CNNs, suggesting that the learned classifiers are much more homogenous.}
    \label{fig:krippendorff}
\end{figure}
We performed experiments with several variants of mixture of experts, finding that overall, it can help, but determining the optimal way to mix LPX domain experts remains challenging. Simple averaging of domain experts (\S\ref{sec:avg}) gives better performance on both sentiment analysis and rumour detection over the single model baseline. Learned attention (\S\ref{sec:attention}) has a net positive effect on performance for sentiment analysis and a negative effect for rumour detection compared to the single model baseline. Additionally, simple averaging of domain experts consistently outperforms a learned sample by sample attention. This highlights the difficulty in utilizing large pretrained transformer models to learn to attend to the predictions of domain experts.

\paragraph{Comparing agreement} To provide some potential explanation for why it is difficult to learn to attend to domain experts, we compare the agreement on the predictions of domain experts of one of our models based on DistilBert, versus a model based on CNNs (\autoref{fig:krippendorff}). CNN models are chosen in order to compare the agreement using our approach with an approach which has been shown to work well with mixture of experts on this data~\cite{guo2018multi}. Each CNN consists of an embedding layer initialized with 300 dimensional FastText embeddings~\cite{bojanowski2017enriching}, a series of 100 dimensional convolutional layers with widths 2, 4, and 5, and a classifier. The end performance is on par with previous work using CNNs~\cite{li2018s} (78.8 macro averaged accuracy, validation accuracies of the individual models are between 80.0 and 87.0). Agreement is measured using Krippendorff's alpha~\cite{krippendorff2011computing} between the predictions of domain experts on test data. 

We observe that the agreement between DistilBert domain experts on test data is significantly higher than that of CNN domain experts, indicating that the learned classifiers of each expert are much more similar in the case of DistilBert. Therefore, it will potentially be more difficult for a mixing function on top of DistilBert domain experts to gain much beyond simple averaging, while with CNN domain experts, there is more to be gained from mixing their predictions. This effect may arise because each DistilBert model is highly pre-trained already, hence there is little change in the final representations, and therefore similar classifiers are learned between each domain expert.
\section{Conclusion}
In this work, we investigated the problem of multi-source domain adaptation with large pretrained transformer models. Both domain adversarial training and mixture of experts techniques were explored. While domain adversarial training could effectively induce more domain agnostic representations, it had a mixed effect on model performance. Additionally, we demonstrated that while techniques for mixing domain experts can lead to improved performance for both sentiment analysis and rumour detection, determining a beneficial mixing of such experts is challenging. The best method we tested was a simple averaging of the domain experts, and we provided some evidence as to why this effect was observed. We find that LPX models may be better suited for data-driven techniques such as that of~\citet{gururangan2020don}, which focus on inducing a better prior into the model through pretraining, as opposed to techniques which focus on learning a better posterior with architectural enhancements. We hope that this work can help inform researchers of considerations to make when using LPX models in the presence of domain shift.

\section*{Acknowledgements}
$\begin{array}{l}\includegraphics[width=1cm]{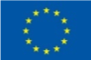} \end{array}$ This project has received funding from the European Union's Horizon 2020 research and innovation programme under the Marie Sk\l{}odowska-Curie grant agreement No 801199. 



\bibliography{anthology,acl2020}
\bibliographystyle{acl_natbib}

\clearpage

\appendix
\section{BERT Domain Adversarial Training Results}
\label{sec:bert_appendix}
Additional results on domain adversarial training with Bert can be found in \autoref{tab:bert_appendix_results}.

\begin{table*}[t!]
    \centering
    \fontsize{10}{10}\selectfont
    \begin{tabular}{l c c c c c | c c c c c c c }
    \toprule 
     Method &\multicolumn{5}{c}{Sentiment Analysis (Accuracy)}&\multicolumn{6}{c}{Rumour Detection (F1)}\\
    \cmidrule(lr){2-6}
    \cmidrule(lr){7-12}
     & D & E & K & B & macroA & CH & F & GW & OS & S & $\mu$F1\\
    \midrule
        Bert & 90.3& 91.6& 91.7& 90.4& 91.0& 66.4& 46.2& 68.3& 67.3& 62.3& 63.3\\
        \midrule
        Bert-Adv-12 & 89.8& 91.4& 91.2& 90.1& 90.6 & 66.6& 47.8& 62.5& 65.3& 62.8& 62.5\\
        Bert-Adv-4 & 89.9& 91.1& 91.7& 90.4& 90.8& 65.6& 43.6& 71.0& 68.1& 60.8& 62.8\\
    \bottomrule 

    \end{tabular}
    \caption{Experiments for sentiment analysis in (D)VD, (E)lectronics, (K)itchen and housewares, and (B)ooks domains and rumour detection for different events ((C)harlie(H)ebdo, (F)erguson, (G)erman(W)ings, (O)ttawa(S)hooting, and (S)ydneySiege) using leave-one-out cross validation for BERT. Results are averaged across 3 random seeds. The results for sentiments analysis are in terms of accuracy and the results for rumour detection are in terms of F1.}
    \label{tab:bert_appendix_results}
\end{table*}

\section{Reproducibility}

\subsection{Computing Infrastructure}
All experiments were run on a shared cluster. Requested jobs consisted of 16GB of RAM and 4 Intel Xeon Silver 4110 CPUs. We used a single NVIDIA Titan X GPU with 12GB of RAM.

\subsection{Average Runtimes}
The average runtime performance of each model is given in \autoref{tab:runtimes}. Note that different runs may have been placed on different nodes within a shared cluster, thus why large time differences occurred. 
\begin{table*}
    \centering
    \fontsize{10}{10}\selectfont
    \begin{tabular}{l c c}
    \toprule 
     Method &Sentiment Analysis &Rumour Detection\\
    \midrule
        Basic & 0h44m37s& 0h23m52s\\
        \midrule
        Adv-6 & 0h54m53s& 0h59m31s\\
        Adv-3 & 0h53m43s& 0h57m29s\\
        \midrule
        Independent-Avg & 1h39m13s& 1h19m27\\
        Independent-Ft & 1h58m55s& 1h43m13\\
        MoE-Avg & 2h48m23s& 4h03m46s\\
        MoE-Att & 2h49m44s& 4h07m3s\\
        MoE-Att-Adv-6 & 4h51m38s& 4h58m33s\\
        MoE-Att-Adv-3 & 4h50m13s& 4h54m56s\\
        MoE-DC & 3h23m46s& 4h09m51s\\
    \bottomrule 

    \end{tabular}
    \caption{Average runtimes for each model on each dataset (runtimes are taken for the entire run of an experiment).}
    \label{tab:runtimes}
\end{table*}

\subsection{Number of Parameters per Model}
The number of parameters in each model is given in \autoref{tab:num_params}.
\begin{table*}
    \centering
    \fontsize{10}{10}\selectfont
    \begin{tabular}{l c c}
    \toprule 
     Method &Sentiment Analysis &Rumour Detection\\
    \midrule
        Basic & 66,955,010& 66,955,010\\
        \midrule
        Adv-6 & 66,958,082& 66,958,850\\
        Adv-3 & 66,958,082& 66,958,850\\
        \midrule
        Independent-Avg & 267,820,040& 334,775,050\\
        Independent-Ft & 267,820,040& 334,775,050\\
        MoE-Avg & 267,820,040& 334,775,050\\
        MoE-Att & 268,999,688& 335,954,698\\
        MoE-Att-Adv-6 & 269,002,760& 335,958,538\\
        MoE-Att-Adv-3 & 269,002,760& 335,958,538\\
        MoE-DC & 267,821,576& 334,777,354\\
    \bottomrule 

    \end{tabular}
    \caption{Number of parameters in each model}
    \label{tab:num_params}
\end{table*}

\subsection{Validation Performance}
The validation performance of each tested model is given in \autoref{tab:val_performance}.
\begin{table*}
    \centering
    \fontsize{10}{10}\selectfont
    \begin{tabular}{l c c}
    \toprule 
     Method &Sentiment Analysis (Acc) &Rumour Detection (F1)\\
    \midrule
        Basic & 91.7& 82.4\\
        \midrule
        Adv-6 & 91.5& 83.3\\
        Adv-3 & 91.2& 83.4\\
        \midrule
        Independent-Avg & 92.7& 82.8\\
        Independent-Ft & 92.6& 82.5\\
        MoE-Avg & 92.2& 83.5\\
        MoE-Att & 92.0& 83.3\\
        MoE-Att-Adv-6 & 91.2& 83.3\\
        MoE-Att-Adv-3 & 91.4& 82.8\\
        MoE-DC & 89.8& 84.6\\
    \bottomrule 

    \end{tabular}
    \caption{Average validation performance for each of the models on both datasets.}
    \label{tab:val_performance}
\end{table*}

\subsection{Evaluation Metrics}
The primary evaluation metrics used were accuracy and F1 score. For accuracy, we used our implementation provided with the code. The basic implementation is as follows.
\begin{equation*}
    \text{accuracy} = \frac{tp + tn}{tp+fp+tn+fn}
\end{equation*}
We used the sklearn implementation of \texttt{precision\_recall\_fscore\_support} for F1 score, which can be found here: \url{https://scikit-learn.org/stable/modules/generated/sklearn.metrics.precision_recall_fscore_support.html}. Briefly:
\begin{equation*}
   p = \frac{tp}{tp + fp} 
\end{equation*}
\begin{equation*}
   r = \frac{tp}{tp + fn} 
\end{equation*}
\begin{equation*}
   F1 = \frac{2*p*r}{p+r} 
\end{equation*}
where $tp$ are true positives, $fp$ are false positives, and $fn$ are false negatives.

\subsection{Hyperparameters}
We performed and initial hyperparameter search to obtain good hyperparameters that we used across models. The bounds for each hyperparameter was as follows:
\begin{itemize}
    \item Learning rate: [0.00003, 0.00004, 0.00002, 0.00001, 0.00005, 0.0001, 0.001].
    \item Weight decay: [0.0, 0.1, 0.01, 0.005, 0.001, 0.0005, 0.0001].
    \item Epochs: [2, 3, 4, 5, 7, 10].
    \item Warmup steps: [0, 100, 200, 500, 1000, 5000, 10000].
    \item Gradient accumulation: [1,2]
\end{itemize}
We kept the batch size at 8 due to GPU memory constraints and used gradient accumulation instead. We performed a randomized hyperparameter search for 70 trials. Best hyperparameters are chosen based on validation set performance (accuracy for sentiment data, F1 for rumour detection data). The final hyperparameters selected are as follows:
\begin{itemize}
    \item Learning rate: 3e-5.
    \item Weight decay: 0.01.
    \item Epochs: 5.
    \item Warmup steps: 200.
    \item Batch Size: 8
    \item Gradient accumulation: 1
\end{itemize}
Additionally, we set the objective weighting parameters to $\lambda = 0.5$ for the mixture of experts models and $\gamma = 0.003$ for the adversarial models, in line with previous work~\cite{guo2018multi,li2018s}.

\subsection{Links to data}
\begin{itemize}
    \item Amazon Product Reviews~\cite{blitzer2007biographies}:  \url{https://www.cs.jhu.edu/~mdredze/datasets/sentiment/}
    
    \item PHEME~\cite{zubiaga2016analysing}: \url{https://figshare.com/articles/PHEME_dataset_for_Rumour_Detection_and_Veracity_Classification/6392078}.
    
\end{itemize}

\end{document}